\documentclass[conference]{IEEEtran}
\IEEEoverridecommandlockouts
\usepackage{tikz}
\usetikzlibrary{arrows.meta, positioning, shapes.geometric, calc, shadows, patterns}
\usepackage{cite}
\usepackage{subcaption}
\usepackage{amsmath,amssymb,amsfonts}
\usepackage{algorithmic}
\usepackage{graphicx}
\usepackage{textcomp}
\usepackage{xcolor}
\usepackage[colorlinks,linkcolor=red,anchorcolor=red,citecolor=blue]{hyperref}
\usepackage[lined,boxed,ruled]{algorithm2e}

\usepackage{listings}
\usepackage[skins, listings]{tcolorbox}
\newtcblisting{jsonbox}{
    enhanced,  
    colback=gray!5,          
    colframe=gray!60,        
    boxrule=0.1pt,            
    arc=1pt,  
    top=-2mm, 
    bottom=-2mm,
    listing only,             
    listing options={
        language=Java,                    
        basicstyle=\footnotesize\ttfamily, 
        breaklines=true,                  
        columns=fullflexible,              
        showstringspaces=false            
    }
}

\usepackage{paralist}
\graphicspath{{./figs/}}

\def\BibTeX{{\rm B\kern-.05em{\sc i\kern-.025em b}\kern-.08em
    T\kern-.1667em\lower.7ex\hbox{E}\kern-.125emX}}
\begin{document}

\title{GRAIL: A Deep-Granularity Hybrid Resonance Framework for Real-Time Agent Discovery via SLM-Enhanced Indexing}

\author{\IEEEauthorblockN{Jinliang Xu$^{*}$}
\IEEEauthorblockA{\textit{China Academy of Information and Communications Technology}, Beijing, China \\
xujinliang@caict.ac.cn}
}

\maketitle

\begin{abstract}
As the ecosystem of Large Language Model (LLM)-based agents expands rapidly, efficient and accurate Agent Discovery becomes a critical bottleneck for large-scale multi-agent collaboration. Existing approaches typically face a dichotomy: either relying on heavy-weight LLMs for intent parsing, leading to prohibitive latency (often exceeding 30 seconds), or using monolithic vector retrieval that sacrifices semantic precision for speed. To bridge this gap, we propose \textbf{GRAIL} (Granular Resonance-based Agent/AI Link), a novel framework achieving sub-400ms discovery latency without compromising accuracy. GRAIL introduces three key innovations: (1) \textbf{SLM-Enhanced Prediction}, replacing the generalized LLM parser with a specialized, fine-tuned Small Language Model (SLM) for millisecond-level capability tag prediction; (2) \textbf{Pseudo-Document Expansion}, augmenting agent descriptions with synthetic queries to enhance semantic density for robust dense retrieval; and (3) \textbf{MaxSim Resonance}, a fine-grained matching mechanism computing maximum similarity between user queries and discrete agent usage examples, effectively mitigating semantic dilution. Validated on \textbf{AgentTaxo-9K}, our new large-scale dataset of 9,240 agents, GRAIL reduces end-to-end discovery latency by over \textbf{79$\times$} compared to LLM-parsing baselines, while significantly outperforming traditional vector search in Recall@10. This framework offers a scalable, industrial-grade solution for the real-time ``Internet of Agents."
\end{abstract}


\section{Introduction}
\label{sec:introduction}

The rapid evolution of Large Language Models (LLMs) has catalyzed a paradigm shift from isolated conversational bots to interconnected Multi-Agent Systems (MAS) \cite{sapkota2025ai, yang2025agentic}. As envisioned in the concept of the ``Internet of Agents'', thousands of specialized agents—ranging from coding assistants to financial analysts\cite{chiang2024llamp, qiang2023agent}—are being deployed to collaboratively solve complex tasks. In this ecosystem, \textit{Agent Discovery}, the process of accurately matching a user's natural language intent to the most capable agent among a massive candidate pool, has emerged as a critical infrastructure challenge\cite{yang2023hierarchical, rodriguez2023approach}.

Despite the proliferation of agent protocols \cite{yang2025survey}, achieving real-time, high-precision agent discovery remains an unsolved problem. Existing approaches predominantly fall into two categories, each with significant limitations:

\textbf{1. LLM-Centric Parsing (High Latency):} Some frameworks rely on general-purpose LLMs (e.g., Gemini 3, DeepSeek) to parse user intent and generate structured queries (e.g., SQL or filtered JSON) before searching the database. While highly accurate in understanding complex intent, this ``Think-then-Lookup'' paradigm incurs prohibitive latency \cite{mei2025omnirouter}. Our empirical analysis shows that the LLM parsing stage alone often consumes over 30 seconds per query, rendering it unsuitable for real-time interaction or high-frequency agent-to-agent negotiation.

\textbf{2. Monolithic Dense Retrieval (Low Precision):} Alternatively, lightweight approaches utilize vector databases to perform dense retrieval on agent descriptions. While efficient (achieving millisecond-level response), these methods typically compress an agent's diverse capabilities (descriptions, tags, and usage examples) into a single embedding vector. This ``coarse-grained'' compression leads to semantic drift, where subtle user constraints (e.g., specific API version or pricing model) are lost, resulting in the retrieval of irrelevant agents.

\begin{figure}[!ht] 
    \centering
    \includegraphics[width=3.5in]{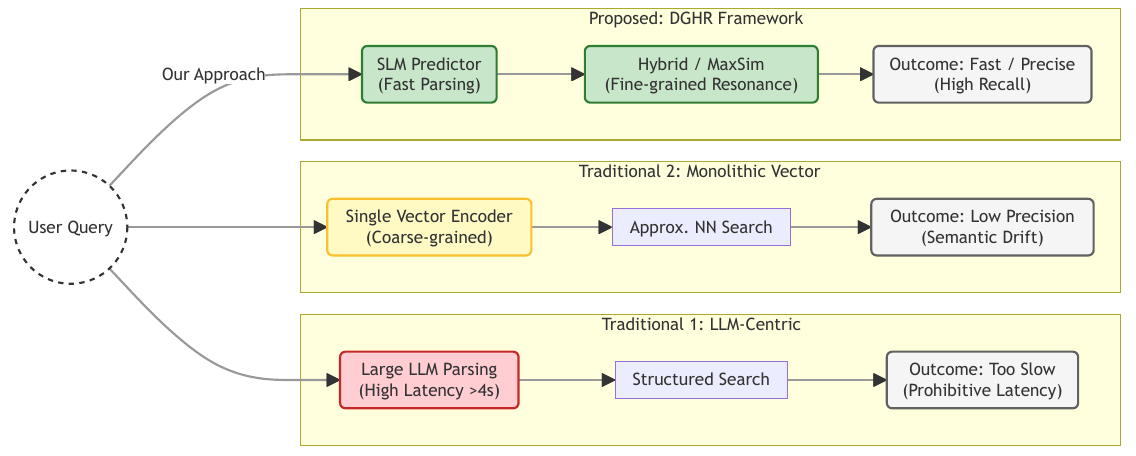} 
    \caption{Comparison of Agent Discovery Paradigms.
\textit{LLM-Centric Parsing} suffers from high latency ($>30$s) due to heavy inference. 
    \textit{Monolithic Vector Retrieval} sacrifices semantic precision for speed, leading to low recall. 
    Our proposed \textbf{GRAIL Framework} leverages a specialized SLM predictor and fine-grained MaxSim resonance to achieve sub-400ms latency without compromising accuracy.}
    \label{fig:comparison}
\end{figure}

To bridge this gap between the precision of LLM reasoning and the speed of vector search, we propose \textbf{GRAIL} (Granular Resonance-based Agent/AI Link), a framework designed for real-time (sub-400ms) discovery of agents at scale. As shown in Fig. \ref{fig:comparison}, unlike previous methods that treat agent metadata as a monolithic text block, GRAIL decouples the discovery process into three granular dimensions: \textit{Tags}, \textit{Context}, and \textit{Intent}. 
Instead of relying on heavy-weight LLMs for online parsing, GRAIL employs a fine-tuned Small Language Model (SLM, $~3$B parameters) to predict capability tags in milliseconds, enabling rapid candidate filtering. Furthermore, we introduce a \textit{MaxSim Resonance} mechanism that computes the interaction between the user query and discrete agent usage examples (shots), ensuring precise intent matching without the computational cost of full-text cross-encoders.

\begin{itemize}
    \item \textbf{We propose GRAIL, a SLM-driven, deep-granularity framework for large-scale agent discovery that targets the latency--precision trade-off.} By replacing heavy-weight LLM parsing with a compact SLM and a streamlined retrieval pipeline, GRAIL achieves sub-400ms end-to-end latency while improving top-$K$ accuracy over LLM-centric routing and standard dense retrieval.

    \item \textbf{We design a Deep-Granularity Hybrid Indexing scheme that separates agent signals into Tag, Context, and Intent dimensions.} By combining SLM-predicted sparse tags with Pseudo-Document--expanded dense descriptions, it enables robust hybrid recall and reduces vocabulary mismatch between metadata and user queries.

    \item \textbf{We develop MaxSim Resonance, a fine-grained matching mechanism over agent usage-example matrices instead of monolithic embeddings.} Late-interaction style maximum similarity between queries and discrete capability modes mitigates semantic dilution in multi-functional agents and yields clear gains in Recall@$K$ and MRR.

    \item \textbf{We release AgentTaxo-9K, a hierarchically structured benchmark for agent discovery with 9{,}240 agents and 27{,}720 queries, and use it to evaluate GRAIL.} Based on an industry--subdomain--function taxonomy and diverse query intents, AgentTaxo-9K provides a standardized testbed where GRAIL consistently outperforms strong baselines in effectiveness and efficiency.
\end{itemize}

\section{Related Work}

Our work sits at the intersection of agent interconnection protocols, efficient neural routing, and fine-grained information retrieval.
In this section, we review the existing landscape and position GRAIL as a complementary indexing layer for the emerging ``Internet of Agents" \cite{dhanasekar2025survey}.

\subsection{Agent Interconnection Protocols \& Discovery Standards}
The infrastructure for the Internet of Agents is being shaped by a series of emerging communication protocols that standardize agent identity and capability description \cite{ehtesham2025survey, liu2025acps}.

\textbf{1. Agent Card-Based Protocols (e.g., Google A2A):} 
Recent industry initiatives, such as the \textbf{Google A2A (Agent-to-Agent)} protocol, utilize \textit{Agent Cards} as standardized manifests to encapsulate agent identities, capabilities, and endpoints \cite{surapaneni2025announcing, huang2025novel}.
While Agent Cards provide a unified metadata schema for interoperability, existing implementations often rely on static registries or direct handshakes, lacking a high-performance engine for fuzzy semantic discovery \cite{liao2025agentmaster}.

\textbf{2. Model Context Protocol (MCP):} 
Anthropic's \textbf{Model Context Protocol (MCP)} standardizes the connection between LLMs and external tools, functioning as a ``USB-C for AI'' \cite{anthropic2025introducing}.
However, MCP primarily addresses tool binding within a host's context window.
As the number of tools grows, static context injection becomes infeasible, necessitating active discovery mechanisms like ``MCP-Zero'' \cite{fei2025mcp}.

\textbf{3. Network-Layer Discovery (e.g., ANP):} 
The \textbf{Agent Network Protocol (ANP)} and similar directory services (e.g., Cisco ADS) address decentralized agent reachability using Distributed Hash Tables (DHTs) or DNS-SD variants \cite{muscariello2025agent, chang2025agent}.
These network-centric approaches excel at routing by ID but struggle with complex, natural language intent matching.
GRAIL complements these protocols by serving as a dynamic, semantic indexing engine that operates on top of standardized metadata (e.g., Agent Cards or MCP manifests).

\subsection{Efficient Routing via Small Language Models}
The paradigm of ``LLM-as-a-Router" has evolved from using giant models to specialized Small Language Models (SLMs).
\begin{inparaenum}
\item {LLM Tool Learning:} 
Pioneering works such as \textbf{Gorilla} \cite{patil2024gorilla} and \textbf{ToolLLM} \cite{qin2023toolllm} have demonstrated that Large Language Models can be fine-tuned to master thousands of APIs.
However, relying on general-purpose LLMs (e.g., Gemini 3) for intent routing introduces significant latency (often $>30$s), creating a bottleneck for real-time agent interaction \cite{greyling_slm_classification_2026,belcak2025small}.
\item {SLM and Knowledge Distillation:} 
To address the latency challenge, recent research advocates for the use of SLMs (parameters $<1$B) as efficient neural routers \cite{belcak2025small}.
\end{inparaenum}
Through \textit{Knowledge Distillation}, capabilities from larger teacher models are transferred to SLMs, enabling them to perform specific intent classification tasks with comparable accuracy but at a fraction of the inference cost (e.g., $<400$ms) \cite{greyling_slm_classification_2026}.
GRAIL adopts this strategy, utilizing a fine-tuned SLM for rapid tag prediction to filter candidates before expensive retrieval operations.

\subsection{Neural Retrieval \& Fine-Grained Late Interaction}
Retrieving the correct agent from a massive pool requires overcoming the ``semantic dilution" problem inherent in traditional dense retrieval.
\textbf{1. Limitations of Monolithic Embeddings:} 
Standard bi-encoder architectures (e.g., DPR) compress an entire document into a single vector \cite{karpukhin-etal-2020-dense, lupart2025disco}.
In the context of multi-functional agents, this compression leads to \textit{context dilution}, where specific capabilities are averaged out, making precise matching difficult \cite{lumer2025tool}.
\textbf{2. Late Interaction (MaxSim):} 
To preserve fine-grained semantic signals, \textbf{ColBERT} introduced the concept of \textit{Late Interaction}, which computes the maximum similarity (MaxSim) between query tokens and document tokens \cite{khattab2020colbert}.
Recent optimizations like \textbf{PLAID} have proven the viability of this approach in low-latency settings \cite{santhanam2022plaidefficientenginelate}.
GRAIL adapts the MaxSim operator from the token level to the ``example level," computing resonance between user queries and discrete agent usage examples.
\textbf{3. Pseudo-Document Expansion (PDE):} 
To bridge the vocabulary mismatch between user queries and technical agent descriptions, generative query expansion methods (e.g., \textbf{Doc2Query}) have been widely adopted \cite{gospodinov2023doc2query}.
GRAIL integrates PDE to augment agent metadata, enhancing the semantic density of the context index.

\section{The GRAIL Framework}
\label{sec:framework}

\begin{figure}[!ht] 
    \centering
    \includegraphics[width=3.5in]{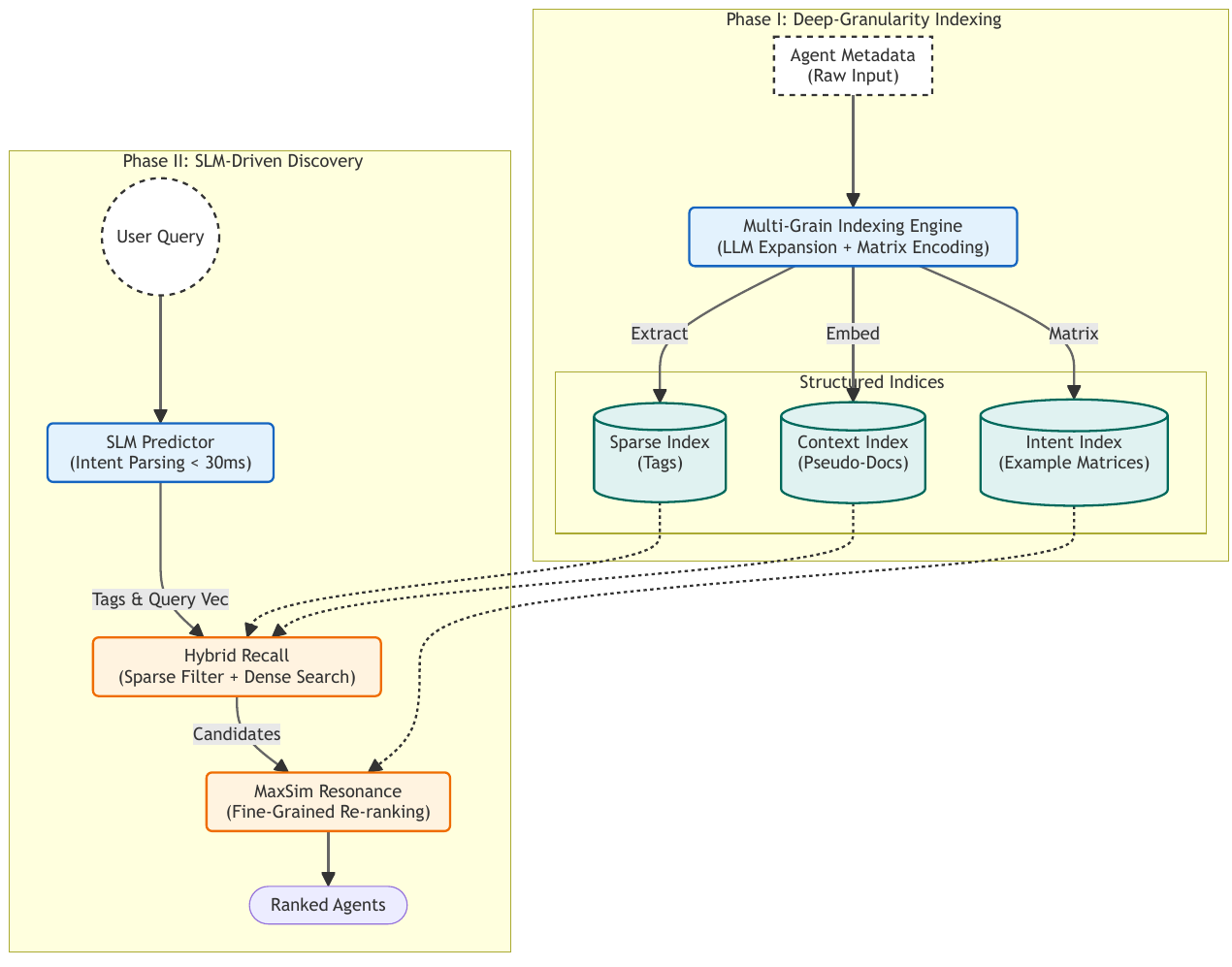} 
    \caption{The Overall Architecture of the GRAIL Framework. }
    \label{fig:architecture}
\end{figure}

In this section, we present the \textbf{GRAIL} framework. We first formulate the agent discovery problem, then detail the offline construction of our tri-dimensional index, and finally describe the online SLM-driven retrieval and MaxSim re-ranking process. Fig. \ref{fig:architecture} illustrates the overall architecture. The system operates in two distinct phases: 
    \textbf{(1) Offline Deep-Granularity Indexing:} Agent metadata is processed into three independent indices---Sparse (Tags), Context (Pseudo-Doc Augmented Vectors), and Intent (Fine-Grained Example Matrices). 
    \textbf{(2) Online SLM-Driven Discovery:} A lightweight SLM predictor ($~3$B params) accelerates intent parsing ($<50ms$), driving a hybrid recall mechanism followed by the MaxSim Resonance re-ranking algorithm to ensure precise intent matching.

\subsection{Problem Formulation}
To illustrate the structure of agents within our framework, consider the following example of an agent's metadata (including \textit{Description}, \textit{Tags} and \textit{Examples}), represented in JSON format:
\begin{jsonbox}
{
    "Name": "HR-Core Automator",
    "Description": "This agent is an expert in optimizing Human Resources workflows within Human Capital Management systems like ...",
    "Tags": ["Security", "Technology", "HR", "HCM", "Workday", ...
    ],
    "Examples": [
        "Draft an API call to Workday to init the onboarding workflow ...",
        "Analyze the attached employee ..."
    ]
}
\end{jsonbox}
Let $\mathcal{A} = \{a_1, a_2, \dots, a_N\}$ denote a repository of $N$ agents. Each agent $a_i$ is defined by a tuple of structured metadata: $a_i = \langle \mathcal{T}_i, \mathcal{D}_i, \mathcal{E}_i \rangle$, where:
\begin{itemize}
    \item $\mathcal{T}_i$ is a set of discrete capability tags (\textit{Tags}), encompassing core domain areas and primary capabilities, alongside both high-frequency and low-frequency yet representative unique tags specific to the agent's functionalities.
    \item $\mathcal{D}_i$ is a natural language description of the agent's functionality (\textit{Description}).
    \item $\mathcal{E}_i = \{e_{i,1}, e_{i,2}, \dots, e_{i,m}\}$ is a list of specific usage examples or few-shot prompts (\textit{Examples}).
\end{itemize}
Given a user query $q$, the objective is to retrieve a ranked subset of agents $\mathcal{R} \subset \mathcal{A}$ ($|\mathcal{R}| \ll N$) such that the relevance score $S(q, a_i)$ is maximized for $a_i \in \mathcal{R}$, subject to a latency constraint $L < 400$ms.

\subsection{Offline: Deep-Granularity Indexing}
Unlike monolithic retrieval systems that compress all agent information into a single vector, GRAIL constructs three distinct indices to capture capability at different granularities.

\subsubsection{Sparse Index via Explicit Tags ($I_{tag}$)}
We construct an inverted index mapping each tag $t \in \bigcup \mathcal{T}_i$ to the list of agents possessing it. This serves as a hard constraint filter. To handle non-explicitly present synonymous tags, we employ a synonym expansion dictionary during index construction.

\subsubsection{Context Index via Pseudo-Document Expansion ($I_{ctx}$)}
Standard dense retrieval on short descriptions ($\mathcal{D}_i$) often suffers from the ``vocabulary mismatch'' problem \cite{weller2024generative}.
We adopt a document expansion strategy. For each agent $a_i$, we utilize a Large Language Model (LLM) to generate a set of potential user queries $\mathcal{Q}^{syn}_i$ based on $\mathcal{D}_i$ \cite{zhang2024query}.
A pseudo-document $\mathcal{D}'_i$ is formed by concatenating the original description and the synthetic queries:
\begin{equation}
    \mathcal{D}'_i = \mathcal{D}_i \oplus \text{Concat}(\mathcal{Q}^{syn}_i)
\end{equation}
We then encode $\mathcal{D}'_i$ into a dense vector $\mathbf{v}_{ctx}^{(i)} \in \mathbb{R}^d$. This index $I_{ctx}$ captures the broad semantic context of the agent.

It is important to note that synthetic queries serve a distinct purpose from the agent's explicit usage examples for several reasons: \begin{inparaenum}
    \item Broader Semantic Coverage: Synthetic queries, generated by an LLM based on the agent's full description, can simulate a wider and more diverse range of user expressions, thus enhancing the broad semantic footprint of the agent for initial dense retrieval.
    \item Addressing Vocabulary Mismatch: They are specifically designed to bridge the lexical gap between formal agent descriptions and varied natural language user queries, optimizing for general recall.
    \item Complementary Role: The usage examples are often limited in number and represent specific, fine-grained interaction patterns. Their role in our framework is to preserve these distinct capability modes for subsequent detailed matching, which is fundamentally different from the broad contextual matching aim of this index.
\end{inparaenum} This strategic separation ensures that each index dimension is optimized for its unique contribution to the multi-stage discovery process.

\subsubsection{Intent Index via Fine-Grained Examples ($I_{int}$)}
This is the core of our deep-granularity approach. Instead of averaging the embeddings of usage examples $\mathcal{E}_i$, which dilutes specific intent signals, we maintain the independence of each example.
For each agent $a_i$, we encode its examples into a matrix of embeddings $\mathbf{M}_i \in \mathbb{R}^{m \times d}$:
\begin{equation}
    \mathbf{M}_i = \left[\text{Enc}(e_{i,1}), \text{Enc}(e_{i,2}), \dots, \text{Enc}(e_{i,m})\right]
\end{equation}
This matrix structure allows the system to preserve distinct capability modes (e.g., an agent capable of both ``coding'' and ``plotting'' will retain strong signals for both, rather than a blurred average).

Notably, the Sparse Index ($I_{tag}$) is a simple inverted index and does not require a vector database. In contrast, the construction of the Context Index ($I_{ctx}$) and Intent Index ($I_{int}$) necessitates the use of a vector database. For this, we leverage a pre-trained sentence transformer like BGE-M3 as the embedding model, which is used directly without further training. The resulting high-dimensional vectors are then organized into the required structures: $I_{ctx}$ stores one vector per agent, representing its overall context, while $I_{int}$ stores a matrix of vectors per agent, where each row corresponds to an encoded usage example. These structured vector data are then persistently stored in specialized vector databases or indexing libraries, enabling subsequent efficient retrieval and computation operations.

\subsection{Online: SLM-Driven Hybrid Retrieval}
To eliminate the latency bottleneck of LLM-based parsing, we introduce a pipelined retrieval mechanism driven by a Small Language Model (SLM).

\subsubsection{SLM Tag Prediction}
Upon receiving a query $q$, we employ a fine-tuned SLM to predict the most likely capability tags $\hat{\mathcal{T}}_q$. The SLM is trained on pairs of $(\text{synthetic query}, \text{tags})$ to minimize the cross-entropy loss. Through this training, the SLM learns to map diverse user intents to relevant tags, regardless of their granularity---encompassing both broad, high-level categories and specific, fine-grained unique tags that reflect specialized agent capabilities.
\begin{equation}
    \hat{\mathcal{T}}_q = \text{SLM}_{\theta}(q)
\end{equation}
Since the output vocabulary is restricted to the system's tag set, inference time is strictly bounded.

\subsubsection{Hybrid Recall}
We perform parallel retrieval to generate a candidate set $\mathcal{C}$:
\begin{itemize}
    \item \textbf{Sparse Path:} Retrieve agents whose capability tag set $\mathcal{T}_i$ contains \textit{any} of the predicted tags in $\hat{\mathcal{T}}_q$. This effectively acts as a broad filter using an OR logic across the predicted tags. Let this set be $\mathcal{C}_{sparse}$.
    \item \textbf{Dense Path:} Encode $q$ into $\mathbf{v}_q$ and perform Approximate Nearest Neighbor (ANN) search on $I_{ctx}$ to retrieve top-$k$ agents. Let this set be $\mathcal{C}_{dense}$.
\end{itemize}
The initial candidate set is the union: $\mathcal{C} = \mathcal{C}_{sparse} \cup \mathcal{C}_{dense}$. This hybrid approach ensures high recall by combining explicit intent (tags) with implicit semantic matching (vectors).

\subsection{MaxSim Resonance Re-ranking}
The candidate set $\mathcal{C}$ is typically small (e.g., $|\mathcal{C}| = 50$). We perform a fine-grained re-ranking using the \textit{MaxSim Resonance} score.
For each candidate agent $a_i \in \mathcal{C}$, we compute the interaction between the query vector $\mathbf{v}_q$ and the agent's example matrix $\mathbf{M}_i$. The resonance score is defined as the maximum cosine similarity found across all examples:
\begin{equation}
    S_{res}(q, a_i) = \max_{j=1}^{m} \left( \frac{\mathbf{v}_q \cdot \mathbf{e}_{i,j}}{\|\mathbf{v}_q\| \|\mathbf{e}_{i,j}\|} \right)
\end{equation}
\textbf{Interpretation:} This operation effectively measures if the user's query ``resonates'' with \textit{any single specific capability} demonstrated in the agent's examples. Unlike average pooling, MaxSim is robust to noise; even if 9 out of 10 examples are irrelevant, a single high-match example triggers a high relevance score.

The final ranking score $S_{final}$ is a weighted sum of the context score and the resonance score:
\begin{equation}
    S_{final}(q, a_i) = \alpha \cdot (\mathbf{v}_q \cdot \mathbf{v}_{ctx}^{(i)}) + (1-\alpha) \cdot S_{res}(q, a_i)
\end{equation}
where $\alpha$ is a hyperparameter balancing broad context context and specific intent. The detailed pseudocode for this entire online discovery pipeline is provided in Algorithm \ref{alg:online_discovery}.

\begin{algorithm}[t]
\caption{GRAIL Online Discovery Process}
\label{alg:online_discovery}
\begin{algorithmic}[1]
\REQUIRE Query $q$, Indices $I_{tag}, I_{ctx}, I_{int}$, SLM model $\mathcal{M}$
\ENSURE Top-$K$ ranked agents $\mathcal{R}$
\STATE \textbf{Step 1: SLM Prediction}
\STATE $\hat{\mathcal{T}}_q \leftarrow \mathcal{M}.\text{predict}(q)$
\STATE $\mathbf{v}_q \leftarrow \text{Encoder}(q)$
\STATE \textbf{Step 2: Hybrid Recall}
\STATE $\mathcal{C}_{sparse} \leftarrow I_{tag}.\text{lookup}(\hat{\mathcal{T}}_q)$
\STATE $\mathcal{C}_{dense} \leftarrow I_{ctx}.\text{search}(\mathbf{v}_q, \text{top\_k}=50)$
\STATE $\mathcal{C} \leftarrow \mathcal{C}_{sparse} \cup \mathcal{C}_{dense}$
\STATE \textbf{Step 3: MaxSim Re-ranking}
\FOR{each agent $a_i$ in $\mathcal{C}$}
    \STATE Load example matrix $\mathbf{M}_i$ from $I_{int}$
    \STATE $S_{res} \leftarrow \max(\mathbf{v}_q \cdot \mathbf{M}_i)$
    \STATE $S_{final} \leftarrow \alpha (\mathbf{v}_q \cdot \mathbf{v}_{ctx}^{(i)}) + (1-\alpha) S_{res}$
\ENDFOR
\STATE Sort $\mathcal{C}$ by $S_{final}$ descending
\RETURN Top-$K$ agents from $\mathcal{C}$
\end{algorithmic}
\end{algorithm}

\begin{figure}[!ht] 
    \centering
    \resizebox{1\linewidth}{!}{
    \begin{tikzpicture}[
        font=\normalsize,
        >=Stealth,
        line width=0.7pt,
        lifeline/.style={draw, dashed, gray!70},
        entity/.style={rectangle, draw, fill=blue!10, minimum width=2.2cm, minimum height=0.8cm, rounded corners=2pt, font=\normalsize},
        engine/.style={rectangle, draw, fill=orange!10, minimum width=2.5cm, minimum height=0.8cm, rounded corners=2pt, font=\normalsize},
        msg/.style={->, thick},
        box/.style={draw, dashed, gray!60, fill=gray!5, fill opacity=0.3, rounded corners=3pt}
    ]

    \node[entity] (user) {User};
    \node[engine, right=1.2cm of user] (eng) {Discovery Engine};
    \node[entity, right=1.2cm of eng] (models) {SLM / BGE};
    \node[entity, right=1.2cm of models] (db) {Hybrid Indices};

    \foreach \x in {user, eng, models, db} {
        \draw[lifeline] (\x.south) -- +(0,-9.6);
    }


    \draw[msg] ($(user.south)+(0,-0.5)$) -- node[above] {Query $q$} ($(eng.south)+(0,-0.5)$);

    \node[anchor=west, blue!70!black, font=\small\bfseries] at ($(eng.south)+(0.1,-0.9)$) {Step 1: Intent Parsing};
    \draw[msg] ($(eng.south)+(0,-1.5)$) -- node[above] {Infer($q$)} ($(models.south)+(0,-1.5)$);
    \draw[msg, dashed] ($(models.south)+(0,-2.2)$) -- node[above] {Tags $\hat{\mathcal{T}}_q$, Vector $\mathbf{v}_q$} ($(eng.south)+(0,-2.2)$);

    \node[anchor=west, blue!70!black, font=\small\bfseries] at ($(eng.south)+(0.1,-3.0)$) {Step 2: Hybrid Recall};
    
    \draw[box] ($(eng.south)+(-0.3,-3.3)$) rectangle ($(db.south)+(0.5,-5.9)$);
    \node[anchor=north west, black, font=\small\itshape] at ($(eng.south)+(-0.3,-3.3)$) {Parallel Path};

    \draw[msg] ($(eng.south)+(0,-4.0)$) -- node[above, font=\small] {Lookup $\hat{\mathcal{T}}_q$ in $I_{tag}$} ($(db.south)+(0,-4.0)$);
    \draw[msg, dashed] ($(db.south)+(0,-4.5)$) -- node[above, font=\small] {$\mathcal{C}_{sparse}$} ($(eng.south)+(0,-4.5)$);
    
    \draw[msg] ($(eng.south)+(0,-5.0)$) -- node[above, font=\small] {Search $\mathbf{v}_q$ in $I_{ctx}$} ($(db.south)+(0,-5.0)$);
    \draw[msg, dashed] ($(db.south)+(0,-5.7)$) -- node[above, font=\small] {$\mathcal{C}_{dense}$} ($(eng.south)+(0,-5.7)$);

    \node[anchor=west, blue!70!black, font=\small\bfseries] at ($(eng.south)+(0.1,-6.5)$) {Step 3: Fusion \& MaxSim Resonance};
    
    \node[draw, fill=white, inner sep=2pt, font=\small] (merge) at ($(eng.south)+(0,-7.0)$) {$\mathcal{C} = \mathcal{C}_{sparse} \cup \mathcal{C}_{dense}$};
    
    \draw[msg] ($(eng.south)+(0,-7.7)$) -- node[above] {Fetch $\mathbf{M}_i$ for $\mathcal{C}$ from $I_{int}$} ($(db.south)+(0,-7.7)$);
    \draw[msg, dashed] ($(db.south)+(0,-8.2)$) -- node[above] {Matrices $\{\mathbf{M}_i\}$} ($(eng.south)+(0,-8.2)$);

    \node[draw, fill=white, inner sep=2pt, font=\small] (maxsim) at ($(eng.south)+(0,-8.6)$) {Compute $S_{res}(q, a_i)$};
    \draw[msg] ($(eng.south)+(0,-9.4)$) -- node[above] {Top-$K$ Ranked Agents} ($(user.south)+(0,-9.4)$);

    \end{tikzpicture}
    }
    \caption{Workflow of the GRAIL Online Discovery Process.}
    \label{fig:workflow}
\end{figure}
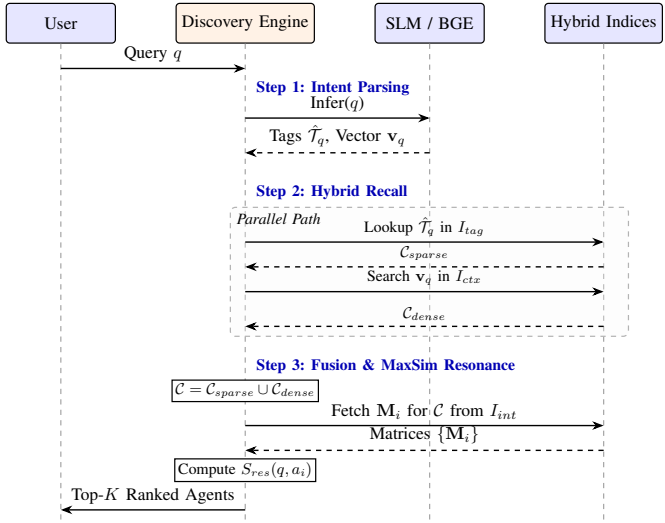

The workflow of the GRAIL online discovery process is as shown in Fig. \ref{fig:workflow}. The Discovery Engine coordinates the intent parsing (SLM/BGE), performs parallel hybrid recall across sparse ($I_{tag}$) and dense ($I_{ctx}$) indices, and executes MaxSim resonance re-ranking using fine-grained example matrices ($I_{int}$).

\subsection{Theoretical Analysis: The Signal Dilution Problem}
To rigorously justify the necessity of the MaxSim Resonance mechanism, we analyze the behavior of relevance scores under the \textit{multi-skilled agent assumption}.

\textbf{Theorem 1 (Semantic Dilution in Mean Pooling).} \textit{Let an agent $a$ possess $m$ distinct capabilities represented by normalized embedding vectors $\mathcal{E} = \{\mathbf{e}_1, \dots, \mathbf{e}_m\}$. Let $q$ be a user query perfectly aligned with a specific capability $\mathbf{e}_k$ (i.e., $q \cdot \mathbf{e}_k = 1$). If the agent's capabilities are distinct (orthogonal in high-dimensional space), the relevance score retrieved by monolithic mean-pooling decays linearly with $m$, whereas MaxSim preserves the unitary signal strength.}

\begin{IEEEproof}
Let $\mathbf{v}_{avg}$ be the monolithic vector representation of agent $a$ used in traditional dense retrieval, typically computed via mean pooling:
\begin{equation}
    \mathbf{v}_{avg} = \frac{1}{m} \sum_{i=1}^{m} \mathbf{e}_i
\end{equation}
Assume the embedding space is high-dimensional (e.g., $d=768$) such that distinct capabilities are approximately orthogonal. That is, for $i \neq j$, $\mathbf{e}_i \cdot \mathbf{e}_j \approx \epsilon$, where $\epsilon \to 0$.

\textbf{Case 1: Traditional Mean Pooling Score ($S_{avg}$).}
The relevance score for query $q = \mathbf{e}_k$ is:
\begin{align}
    S_{avg} &= q \cdot \mathbf{v}_{avg} = \mathbf{e}_k \cdot \left( \frac{1}{m} \sum_{i=1}^{m} \mathbf{e}_i \right) \\
    &= \frac{1}{m} \left( \mathbf{e}_k \cdot \mathbf{e}_k + \sum_{i \neq k} \mathbf{e}_k \cdot \mathbf{e}_i \right) \\
    &\approx \frac{1}{m} (1 + (m-1)\epsilon) \approx \frac{1}{m}
\end{align}
As the number of capabilities $m$ increases (e.g., a ``Swiss Army Knife'' agent), the retrieval score $S_{avg}$ asymptotically approaches 0. This phenomenon, which we term \textit{Semantic Dilution}, causes capable agents to be buried in the ranking list by specialized but irrelevant agents.

\textbf{Case 2: GRAIL MaxSim Score ($S_{max}$).}
In our framework, the relevance score is determined by the resonance mechanism:
\begin{equation}
    S_{max} = \max_{i} (q \cdot \mathbf{e}_i) = \max (\mathbf{e}_k \cdot \mathbf{e}_k, \dots) = 1
\end{equation}
\textbf{Conclusion:} $S_{max} \gg S_{avg}$ for any $m > 1$.
Thus, GRAIL maintains a constant high-confidence score invariant to the diversity of the agent's other capabilities, theoretically guaranteeing higher Recall@K for multi-purpose agents.
\end{IEEEproof}

\section{The AgentTaxo-9K Benchmark}
\label{sec:benchmark}

To evaluate the discovery performance of GRAIL in a massive and complex ecosystem, we establish \textbf{AgentTaxo-9K}\footnote{https://github.com/wolfbrother/AgentTaxo-9K}, a novel large-scale benchmark dataset. Unlike existing datasets that often suffer from limited domain coverage or flat structures, AgentTaxo-9K provides a fine-grained, hierarchical organization of agent metadata and high-quality evaluation queries.

\subsection{Dataset Generation and Rationality}
The construction of AgentTaxo-9K follows a rigorous top-down generation pipeline designed to ensure structural integrity, functional diversity, and semantic richness. The process is governed by the following key rules:

\begin{itemize}
    \item \textbf{Industry Anchoring via OASF:} The taxonomy is anchored by $154$ professional industry categories derived from the \textit{Open Agentic Schema Framework (OASF)}\footnote{https://github.com/agntcy/oasf}. This ensures that the dataset's root nodes align with established industrial and service sectors.
    
    \item \textbf{MECE-based Sub-domain Expansion:} Each industry is expanded into $6$ distinct sub-domains. To ensure the rationality of this expansion, we enforce the \textbf{MECE (Mutually Exclusive and Collectively Exhaustive)} principle. This rule ensures that the sub-domains within an industry cover its functional breadth without semantic overlap, facilitating clear categorical boundaries.
    
    \item \textbf{Functional Niche Specialization:} For each sub-domain, $10$ unique agents are populated. The generation rules require that every agent must possess a distinct functional niche. Metadata—including a professional Name, a $50$-$100$ word Description, and a set of $5$-$8$ tags—is meticulously crafted to highlight the agent's core capabilities and unique features that distinguish it from its peers in the same sub-domain.
    
    \item \textbf{Tagging Hierarchy:} Each agent's tag list is structured hierarchically to manifest its identity: $1$ industry-level tag, $3$ sub-domain-specific tags, and $1$-$4$ unique functional tags. This cumulative tagging strategy provides the multi-level granularity required for precise capability-based discovery.
    
    \item \textbf{Multi-Intent Query Derivation:} To evaluate the discovery engine's robustness against varied user behaviors, three distinct types of queries are generated for each agent:
    (1) \textbf{Capability-based:} Direct inquiries regarding specific functions;
    (2) \textbf{Scenario-based:} Descriptions of real-world business problems or pain points;
    (3) \textbf{Keyword-style:} Concise, search-engine-style phrases focusing on niche specializations. 
\end{itemize}

\subsection{Dataset Statistics}
AgentTaxo-9K represents one of the most comprehensive benchmarks for agent discovery to date. The hierarchical statistics of the dataset are summarized as follows:

\begin{itemize}
    \item \textbf{Scale:} The dataset comprises $154$ industries, which expand into $924$ sub-domains. With $10$ agents per sub-domain, the benchmark contains a total of \textbf{9,240} unique agents.
    
    \item \textbf{Metadata Density:} Every agent entry is equipped with a complete metadata profile, including a structured name, a detailed technical description, and a set of $5$-$8$ descriptive tags. Additionally, each agent is accompanied by $3$ high-quality usage examples to ground its functional scope.
    
    \item \textbf{Query Volume:} The evaluation set consists of \textbf{27,720} distinct user queries ($3$ queries per agent), covering a wide spectrum of linguistic patterns and search intents.
    
\end{itemize}

\begin{figure}[!ht]
    \centering
    \begin{subfigure}[b]{3in}
        \centering
        \includegraphics[width=\textwidth]{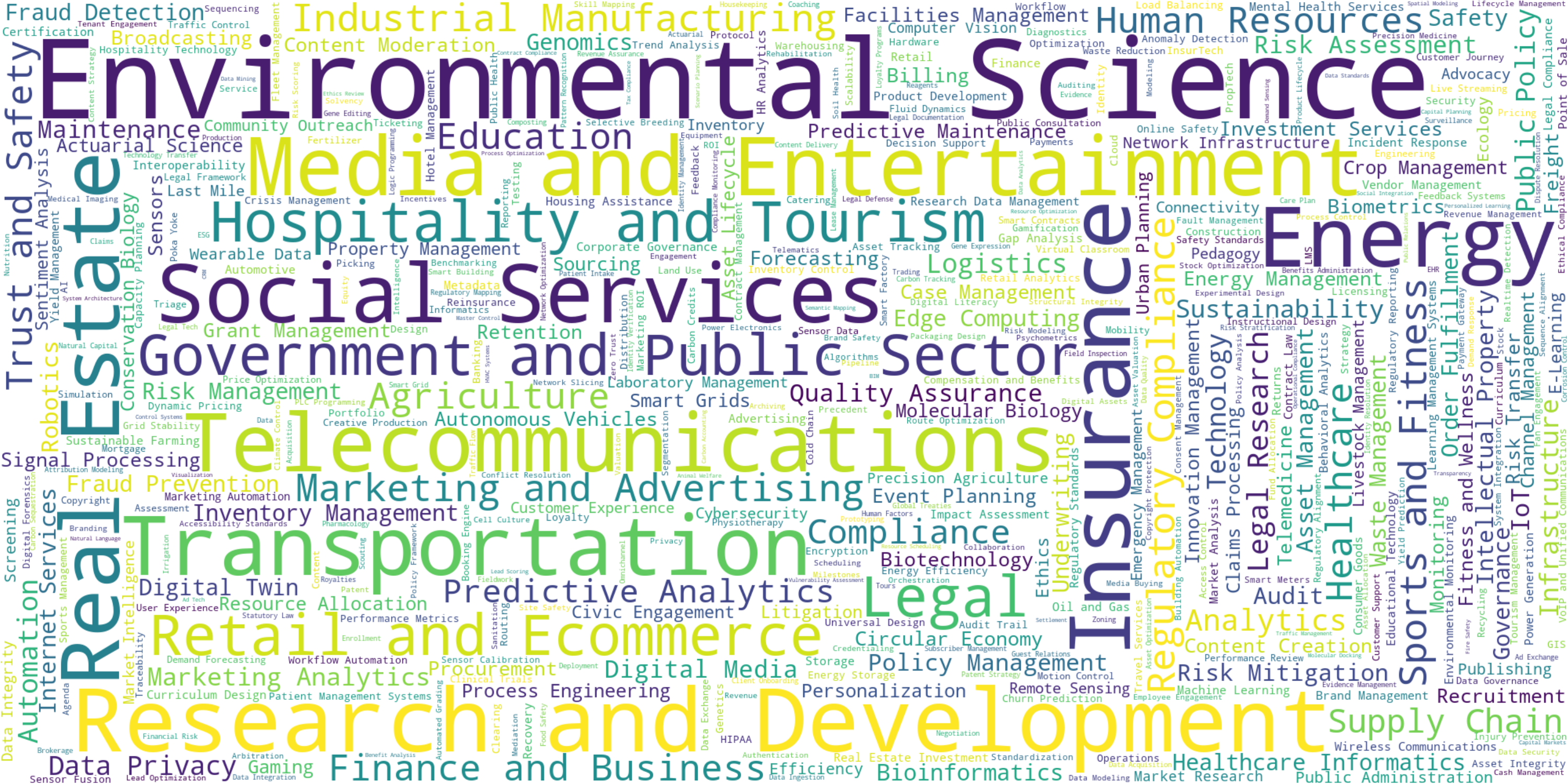}
        \caption{Word cloud of high-frequency tags.}
        \label{fig:wordcloud}
    \end{subfigure}
    \par\vspace{1em}
    \begin{subfigure}[b]{3.5in}
        \centering
        \includegraphics[width=0.8\textwidth]{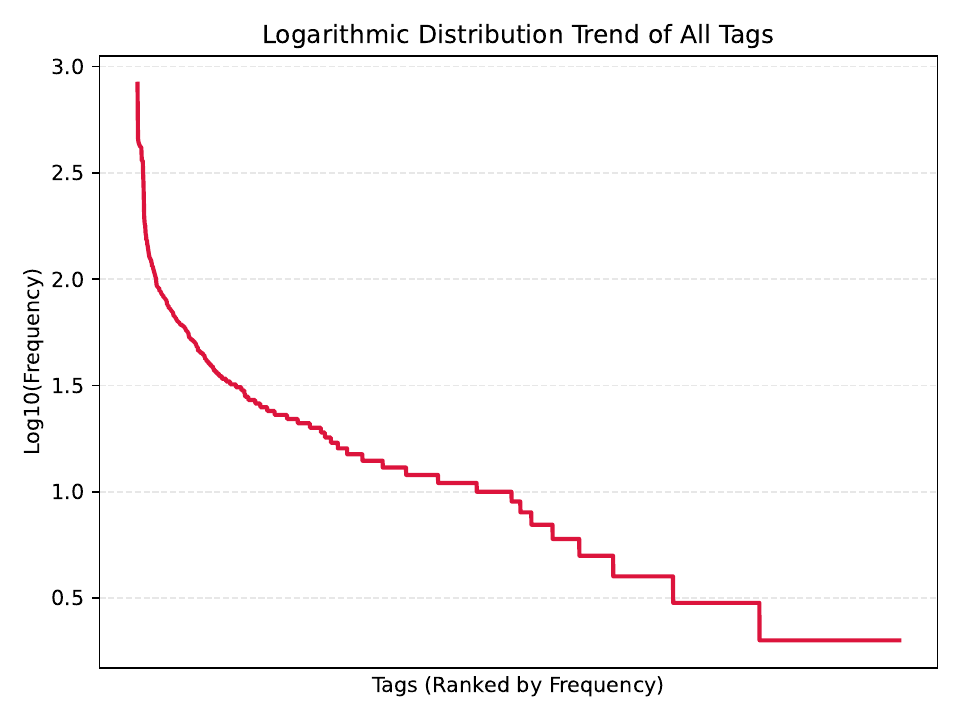}
        \caption{Log-transformed frequency distribution trend of all tags.}
        \label{fig:trendline}
    \end{subfigure}
    \caption{Multi-dimensional distribution analysis of tags (Top: Tag word cloud; Bottom: Log-transformed frequency trend)}
    \label{fig:tag_analysis}
\end{figure}

By leveraging the OASF-based taxonomy and MECE expansion, AgentTaxo-9K spans highly diverse fields—ranging from \textit{Environmental Science} and \textit{Quantum Computing} to \textit{Regulatory Compliance} and \textit{Sports Management}—providing a rigorous testbed for semantic routing and discovery algorithms. Statistical analysis of the 3,194 unique tags within the dataset reveals a characteristic long-tail distribution (Fig. \ref{fig:trendline}), where foundational sectors such as \textit{Environmental Science} and \textit{Insurance}  establish a robust structural core. As visualized in the word cloud (Fig. \ref{fig:wordcloud}), this taxonomy effectively captures prominent industrial domains while maintaining a granular reach into highly specialized niche areas, including Quantum Computing and Bioinformatics. The log-transformed frequency curve underscores the dataset's extensive coverage, ensuring that discovery algorithms are evaluated not only on high-frequency general requests but also on the sparse, complex edge cases essential for cross-domain agent interoperability.

\section{Experimental Evaluation}
\label{sec:experiments}

To validate the effectiveness of the GRAIL framework, we conducted extensive experiments focusing on two key dimensions: \textit{Retrieval Accuracy} (Recall and MRR) and \textit{System Latency}. 
In this section, we describe the  experimental setup, baseline methods, and the quantitative results.

\subsection{Experimental Setup}

\subsubsection{Hardware \& Software Environment}
All experiments were conducted on a Linux server running \textbf{Ubuntu 22.04 LTS}. The hardware configuration consists of a single \textbf{NVIDIA T4 GPU} (15.0 GB VRAM), \textbf{12.7 GB System RAM}, and 112 GB Disk storage. The framework was implemented using \textbf{Python 3.12} and \textbf{PyTorch 2.1}.
Vector operations and MaxSim resonance calculations were accelerated using CUDA tensors to simulate an in-memory high-dimensional retrieval engine.

\subsubsection{Model Configuration}
We strictly decouple the training of the intent parser from the indexing process. The SLM is optimized using a training partition of query-tag pairs, whereas the retrieval engine maintains an exhaustive index of all agents. This \textit{Full-Pool Retrieval} approach rigorously tests the framework's ability to maintain high precision amidst the noise of a massive agent ecosystem. To ensure a rigorous evaluation and avoid data leakage, the synthetic queries used for Pseudo-Document Expansion were generated independently from the ground-truth UserQueries used in the evaluation phase.
\begin{itemize}
    \item \textbf{SLM Predictor (Sparse Path):} We fine-tuned \textbf{Qwen2.5-3B-Instruct} as the intent parser. The model was trained using Low-Rank Adaptation (LoRA) on 27,720 (Query, Tags) pairs derived from the training set. The inference is performed in half-precision (FP16) to minimize latency.
    \item \textbf{Dense Encoder (Context Path):} We utilized \textbf{BGE-Small-EN-v1.5} (33M parameters) as the backbone embedding model, mapping agent descriptions and examples into 384-dimensional vectors.
\end{itemize}

\subsection{Baselines and Comparisons}
We compare GRAIL against three distinct categories of retrieval paradigms to demonstrate the efficacy of our hybrid resonance approach.

\begin{itemize}
    \item \textbf{Baseline 1: Monolithic Dense Retrieval (MDR).} 
    A standard bi-encoder approach where the user query vector is matched against the mean-pooled embedding of the agent's description. This represents the current state-of-the-art in lightweight vector search.
    
    \item \textbf{Baseline 2: Sparse-Only Retrieval (SLM-Sort).} 
    A method relying solely on the SLM to predict tags. Agents possessing the predicted tags are retrieved and ranked randomly. This baseline evaluates the isolated performance of the ``Sparse Semantics".
    
    \item \textbf{Baseline 3: LLM-Centric Routing (LLM).} 
    A ``Think-then-Lookup" approach where a powerful LLM analyzes the query and selects the agent from the full list. While providing an accuracy upper bound, it serves as a baseline for latency comparison.
    
    \item \textbf{Ours: GRAIL (Hybrid + MaxSim).} 
    Our proposed framework integrating SLM-predicted sparse filtering, pseudo-document dense retrieval, and the fine-grained MaxSim Resonance re-ranking on agent examples.
\end{itemize}

Each baseline method employs language models or embedding models in distinct ways. The \textbf{MDR} baseline relies on a pre-trained dense encoder to generate a single, monolithic embedding for each agent's description, against which user queries are matched. For the \textbf{SLM-Sort} baseline, a fine-tuned SLM is utilized to predict capability tags from user queries, enabling lookup in a sparse inverted index. Lastly, the \textbf{LLM} baseline explicitly uses a powerful LLM to directly interpret user intent and route to the most suitable agent.

\subsection{Main Results}

\subsubsection{Retrieval Performance}
Table \ref{tab:main_results} presents the comparative results on AgentTaxo-9K. We report \textbf{Recall@1} (R@1) \textbf{Recall@10} (R@10) and \textbf{Mean Reciprocal Rank} (MRR@10).

\begin{table}[htbp]
\caption{Performance Comparison on AgentTaxo-9K. Best results are bolded.}
\begin{center}
\begin{tabular}{|l|c|c|c|c|}
\hline
\textbf{Method} & \textbf{R@1}& \textbf{R@10} & \textbf{MRR@10} & \textbf{Latency(ms)} \\
\hline
MDR  & 43.94 & 76.77 & 0.5455 & 12.52 \\
SLM-Sort & 0.07 & 0.51 & 0.0017 & 370.18 \\
LLM-Router & 36.84 & 78.95 & 0.5307 & $31186$ \\
\hline
\textbf{GRAIL (Ours)} & \textbf{71.54\%}& \textbf{91.45\%} & \textbf{0.7850} & \textbf{$394.48$} \\
\hline
\end{tabular}
\label{tab:main_results}
\end{center}
\end{table}

As summarized in Table \ref{tab:main_results}, the \textbf{GRAIL} framework demonstrates a decisive advantage over all baselines. It achieves a \textbf{Recall@1 of 71.54\%} and a \textbf{Recall@10 of 91.45\%}, with an \textbf{MRR@10 of 0.7850}. This indicates that GRAIL not only effectively captures the target agent but consistently ranks it as the primary recommendation.

\subsection{Performance and Efficiency Analysis}

\textbf{1) Precision vs. Semantic Dilution:} The performance gap between GRAIL and \textbf{MDR} demonstrates the severity of the \emph{Semantic Dilution} problem. While MDR utilizes dense vectors, its reliance on mean-pooled embeddings causes it to lose functional nuances, resulting in an MRR of only 0.5455. GRAIL's \textbf{MaxSim Resonance} overcomes this by preserving discrete capability signals from usage examples, leading to a 27.6\% absolute improvement in Recall@1.

\textbf{2) The Failure of Sparse-Only Logic:} The near-zero performance of \textbf{SLM-Sort} (Recall@10: 0.51\%) exposes the \emph{One-to-Many Mapping} bottleneck in large-scale repositories. High-cardinality industry tags map to hundreds of potential candidates, rendering sparse-only filtering ineffective without semantic discrimination. This highlights that SLM tag prediction must be paired with deep-granularity indexing to be viable.

\textbf{3) GRAIL vs. LLM-Centric Routing:} The results for \textbf{LLM-Router} reveal a critical trade-off between reasoning depth and practical utility. Despite its high computational cost, the LLM-Router's Recall@1 (36.84\%) is surprisingly lower than even the MDR baseline, likely due to context window noise or \emph{Lost in the Middle} effects when selecting from numerous candidates. Moreover, its staggering latency is nearly \textbf{79$\times$} higher than GRAIL. This underscores that GRAIL achieves superior discovery accuracy while reducing latency by two orders of magnitude compared to heavy-weight LLM parsing. This efficiency confirms that GRAIL can handle complex intent parsing and fine-grained resonance at a negligible computational cost.

\textbf{4) Latency Composition:} To verify the real-time capability, we measured the end-to-end latency per query on the T4 GPU. The average latency of GRAIL is \textbf{$394.48$ ms}, composed of:
\begin{itemize}
    \item SLM Inference: $\approx$ $370.15$ ms
    \item Vector Search \& MaxSim Calculation: $\approx$ $24.33$ ms
\end{itemize}
This result confirms that our pipeline falls within the real-time interaction requirements for agent systems.

\section{Ablation Study}
\label{sec:ablation}

To evaluate the contribution of each component within the GRAIL framework, we conduct an ablation study on the AgentTaxo-9K dataset with four variants: 
(1) \textbf{Full GRAIL} represents the complete system with SLM prediction, hybrid retrieval, and MaxSim resonance; 
(2) \textbf{w/o MaxSim} replaces the fine-grained resonance mechanism with standard mean-pooling cosine similarity to assess the impact of semantic dilution; 
(3) \textbf{w/o SLM} removes the sparse tag filtering, relying solely on dense retrieval to test the robustness of the context index; 
(4) \textbf{MDR} (Monolithic Dense Retrieval) serves as the traditional baseline using only monolithic description embeddings without any GRAIL-specific enhancements.
The quantitative results are summarized in Table \ref{tab:ablation}.

\begin{table}[htbp]
\caption{Ablation Study of GRAIL Components on AgentTaxo-9K.}
\begin{center}
\begin{tabular}{|l|c|c|c|c|}
\hline
\textbf{Variant} & \textbf{R@1} & \textbf{R@10} & \textbf{MRR} & \textbf{Drop (R@10)} \\
\hline
\textbf{Full GRAIL} & \textbf{71.54\%} & \textbf{91.45\%} & \textbf{0.7850} & - \\
\hline
w/o MaxSim & 43.94\% & 76.77\% & 0.5455 & $\downarrow$ 14.68\% \\
w/o SLM & 71.21\% & 90.48\% & 0.7793 & $\downarrow$ 0.97\% \\
MDR & 43.94\% & 76.77\% & 0.5455 & $\downarrow$ 14.68\% \\
\hline
\end{tabular}
\label{tab:ablation}
\end{center}
\end{table}

\subsection{Ablation Analysis}
As shown in Table \ref{tab:ablation}, the \textbf{MaxSim Resonance} mechanism is the most critical contributor to the framework's precision. When MaxSim is replaced by mean pooling (w/o MaxSim), the Recall@10 drops significantly by \textbf{14.68\%}, and the MRR decreases from 0.7850 to 0.5455. This performance collapse empirically validates our \textit{Theorem 1}, demonstrating that monolithic embeddings suffer from severe semantic dilution when representing multi-functional agents. 

Interestingly, the \textbf{SLM Filter} (w/o SLM) shows a marginal drop of only \textbf{0.97\%} in Recall@10. This indicates that our Context Index (augmented by Pseudo-Document Expansion) already provides a robust semantic foundation. However, as noted in our latency analysis, the SLM predictor remains indispensable for achieving sub-50ms intent parsing and providing explicit symbolic alignment for high-precision scenarios. Finally, the identical performance of \textbf{MDR} and the \textbf{w/o MaxSim} variant highlights that without granular resonance, the inclusion of extensive metadata cannot overcome the inherent limitations of coarse-grained vector matching.

\section{Conclusion}

We have presented \textbf{GRAIL}, a novel framework that reconciles the critical conflict between latency and precision in large-scale agent discovery. By orchestrating an SLM-driven intent parser with a fine-grained MaxSim resonance mechanism, GRAIL successfully mitigates the semantic dilution problem inherent in monolithic embeddings while maintaining sub-400ms response times. Our results demonstrate that this deep-granularity approach not only outperforms state-of-the-art baselines in retrieval accuracy but also provides a scalable, high-throughput indexing infrastructure essential for the evolving Internet of Agents ecosystem.

\section*{Acknowledgment}
The authors contributed collectively to the research. However, the specific  visions and conceptual interpretations presented herein are attributed to the corresponding author (xujinliang@caict.ac.cn; jlxufly@gmail.com) to reflect his current stage of thinking, and do not necessarily represent the official views of the other authors or their affiliated institutions.

\bibliographystyle{ieeetr}
\bibliography{RefAbstracts.bib}

\end{document}